\pdfoutput=1
\documentclass[11pt]{article}

\usepackage[preprint]{acl}
\usepackage{times}
\usepackage{latexsym}

\usepackage[T1]{fontenc}
\usepackage[utf8]{inputenc}

\usepackage{microtype}

\usepackage{inconsolata}

\usepackage{graphicx}

\usepackage{booktabs}
\usepackage{amsmath}
\usepackage{multirow}
\usepackage{threeparttable}
\usepackage{xcolor}
\newcommand{\gain}[1]{\textcolor{green!60!black}{#1}}

\usepackage{tabularx,makecell}

\usepackage{siunitx} 
\usepackage{listings}

\newcommand{\pairsplit}{\addlinespace[2pt]}%\cmidrule(lr){1-5}\addlinespace[1pt]}

\title{Scaling, Simplification, and Adaptation: Lessons from Pretraining on Machine-Translated Text}

\author{
    Dan John Velasco$^{\ast}$ \and Matthew Theodore Roque$^{\ast}$ \\
    Samsung R\&D Institute Philippines\\
  \texttt{\{dj.velasco,roque.mt\}@samsung.com} \\
  \small{\textbf{$^{\ast}$Equal Contribution}} \\
  }

\begin{document}
\maketitle
\begin{abstract}

Most languages lack sufficient data for large-scale monolingual pretraining, creating a “data wall.” Multilingual pretraining helps but is limited by language imbalance and the "curse of multilinguality." An alternative is to translate high-resource text with machine translation (MT), which raises three questions: (1) How does MT-derived data scale with model capacity? (2) Can source-side transformations (e.g., simplifying English with an LLM) improve generalization to native text? (3) How well do models pretrained on MT-derived data adapt when continually trained on limited native text? We investigate these questions by translating English into Indonesian and Tamil—two typologically distant, lower-resource languages—and pretraining GPT-2 models (124M–774M) on native or MT-derived corpora from raw and LLM-simplified English. We evaluate cross-entropy loss on native text, along with accuracy on syntactic probes and downstream tasks. Our results show that (1) MT-pretrained models benefit from scaling; (2) source-side simplification harms generalization to native text; and (3) adapting MT-pretrained models on native text often yields better performance than native-only models, even with less native data. However, tasks requiring cultural nuance (e.g., toxicity detection) demand more exposure to native data.
\end{abstract}

% \begin{figure}[t]
%     \centering
%     \includegraphics[width=\columnwidth]{figure_1.pdf}
%     \caption{\textbf{(Top)} Loss vs. model size for Indonesian. \textbf{(Bottom)} Loss vs. model size for Tamil. Loss is evaluated on native text in their respective languages. In both settings, adding more parameters improves loss and models trained on Natural-MT consistently outperform those trained on Simplified-MT.}
%     \label{fig:mt-scaling}
% \end{figure}

\begin{figure}[t]
    \centering
    \includegraphics[width=0.95\columnwidth]{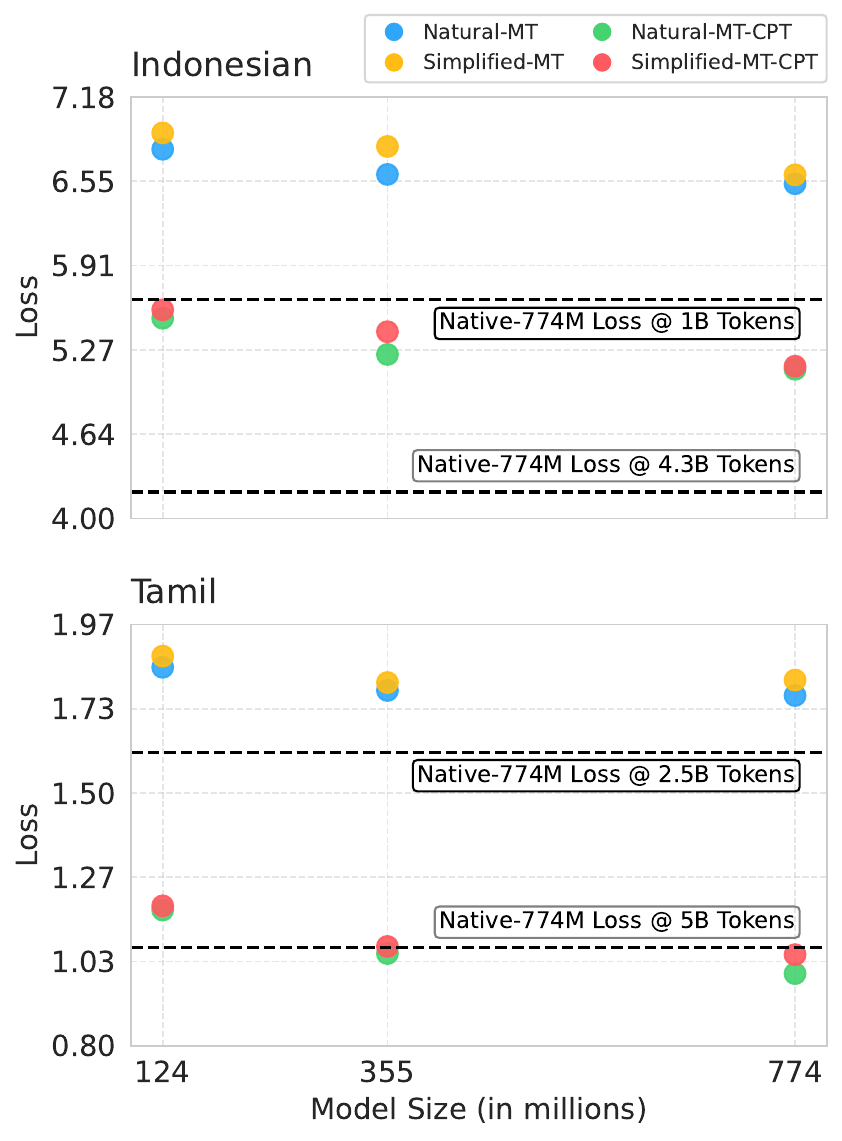}
    \caption{Loss vs. model size for Indonesian \textbf{(top)} and Tamil \textbf{(bottom)}. CPT models are trained with 1B and 2.5B native tokens, respectively. Dashed lines show the loss of the best Native model (Native-774M) as the baseline. Natural-MT outperforms Simplified-MT in both languages. All CPT models exceed Native baselines under equal native token budgets, with Tamil CPT models even surpassing the 5B tokens baseline.}
    \label{fig:cpt}
\end{figure}

\section{Introduction}

Language technologies have advanced rapidly, with Large Language Models (LLMs) achieving strong performance across an array of tasks \cite{NEURIPS2020_1457c0d6, gemmateam2024gemma2improvingopen, qwen2025qwen25technicalreport, grattafiori2024llama3herdmodels}. Scaling studies in pretraining language models show consistent gains with more parameters and more data \cite{kaplan2020scalinglawsneurallanguage, hoffmann2022trainingcomputeoptimallargelanguage}. Yet for most of the world's languages, the native corpora necessary to realize these pretraining benefits are scarce \cite{ustun-etal-2024-aya}, causing models to quickly hit a "data wall"—a performance plateau imposed by limited training data. A common strategy to push past this data wall is multilingual pretraining, which aims to transfer knowledge from high-resource to low-resource languages. However, its effectiveness is constrained by challenges such as language imbalance \citep{chang-etal-2024-multilinguality}, suboptimal multilingual vocabularies \citep{rust-etal-2021-good}, and the “curse of multilinguality” \citep{conneau-etal-2020-unsupervised}.

\pagebreak
One alternative is to translate data from a high-resource language into the target language using machine translation (MT). While this enables large-scale corpus creation, it introduces limitations, including reliance on MT quality and the prevalence of “translationese”—literal phrasing, source-language bias, and cultural mismatches \citep{jalota-etal-2023-translating}. Nonetheless, its scalability makes MT a practical solution to data scarcity. Recent studies investigate the utility of pretraining on MT-derived data (MT pretraining) in both monolingual \cite{doshi-etal-2024-pretraining, alcoba-inciarte-etal-2024-utility} and multilingual settings \citep{wang2025multilinguallanguagemodelpretraining}, consistently reporting downstream performance comparable to models pretrained on native text.

\paragraph{We structure our study around three research questions:}
\begin{enumerate}
  \item[(1)] Does increasing the size of MT-pretrained models improve generalization to native text (cross-entropy loss on held-out native text, syntactic probes, downstream tasks), or does it merely overfit to translation artifacts? %We hypothesize that larger MT-pretrained models will plateau, or even worsen, in perplexity on held-out native text, reflecting overfitting to MT-specific artifacts.
  \item[(2)] Does simplifying source text prior to translation improve the usefulness of MT-derived corpora for pretraining? %We hypothesize that simplifying English before translation will improve translation quality and reduce MT noise, thereby lowering held-out perplexity and improving downstream performance, provided simplification does not substantially reduce lexical or syntactic coverage.
  \item[(3)] Does MT pretraining improve the data efficiency of pretraining on limited native text? %We hypothesize that MT pretraining improves data efficiency and will outperform native-only models even when given fewer native tokens, as measured by perplexity, syntactic probes, and downstream tasks.
\end{enumerate}

\paragraph{Why these questions aren't obvious and why they matter.}
\begin{enumerate}
  \item[(1)]\textbf{Scaling on MT-derived data.} Scaling studies show that performance reliably improves with more parameters and data, but this assumes access to large, high-quality native corpora. When MT-derived data is the only viable option, with its inherent noise and translation artifacts, it remains unclear whether scaling is beneficial or merely leads to overfitting.
  \item[(2)]\textbf{Source-side simplification.} Intuitively, simpler sentences are easier to translate and should yield fewer errors, but at the cost of reduced nuance and lexical/syntactic diversity. If such errors can be reduced in MT-derived data, will this improve pretraining and enhance generalization to native text?
  \item[(3)]\textbf{MT pretraining $\rightarrow$ Native CPT.} MT pretraining may yield transferable features but also embeds translationese patterns that must be unlearned during continual pretraining (CPT) on native text. With a fixed native token budget, is CPT from an MT-pretrained checkpoint more effective than native-only pretraining?
\end{enumerate}

To answer these, we conduct controlled experiments by translating English into Indonesian and Tamil and compare GPT-2 models (124M–774M parameters) pretrained on native corpora against those trained on MT-derived data from both natural and LLM-simplified English sources. We evaluate generalization to native text using cross-entropy loss on held-out data, as well as accuracy on syntactic minimal-pair probes and natural language understanding (NLU) tasks including sentiment analysis (SA), toxicity detection (TD), natural language inference (NLI), and causal reasoning (CR).

\paragraph{Our findings are as follows:}
\begin{itemize}
    \item Scaling MT-pretrained models (124M–774M) improves cross-entropy loss on held-out native text, indicating they do not simply overfit to translation-specific artifacts.
    \item Simplifying source text before translation reduces generalization to native text, likely due to diminished lexical and syntactic variety. Raw translation is therefore both simpler and more effective.
    \item Continual pretraining on limited native text generally improves syntactic probe accuracy and downstream performance, often surpassing native-only models even with less native data. This shows that MT pretraining provides a strong initialization for bootstrapping target-language performance.
    \item MT-pretrained models underperform on tasks requiring cultural nuance, such as toxicity detection, suggesting that such domains demand more extensive native data.
\end{itemize}

To the best of our knowledge, this is the first systematic study of scaling effects in pretraining on MT-derived data, as well as the first exploration of source-side text manipulation prior to translation as a means of enhancing MT data quality.

\pagebreak
\section{Related Work}

\paragraph{Performance gap in low-resource languages.} Recent LLM breakthroughs have centered on high-resource languages like English, where abundant high-quality data is available \cite{joshi-etal-2020-state}. In contrast, low-resource languages still lag due to limited training data and benchmarks. This gap has driven community efforts such as Masakhane \cite{orife2020masakhanemachinetranslation}, SEA-CROWD \cite{lovenia-etal-2024-seacrowd}, and multilingual open-source LLMs like BLOOM \cite{workshop2023bloom176bparameteropenaccessmultilingual} and Aya \cite{ustun-etal-2024-aya}, highlighting the need for data and model development beyond English.

\paragraph{Pretraining on Multilingual Data.}

Multilingual pretraining improves performance in low-resource languages \citep{liu-etal-2020-multilingual-denoising}, offering a path beyond the data wall. Its promise lies in transferring knowledge across languages, but this comes with the “curse of multilinguality” \citep{conneau-etal-2020-unsupervised}, a phenomenon where training on many languages degrades performance on individual languages due to limited capacity and inter-language interference. Despite notable successes \citep{xue-etal-2021-mt5, workshop2023bloom176bparameteropenaccessmultilingual, ustun-etal-2024-aya}, multilingual models still face challenges such as imbalanced data \cite{chang-etal-2024-multilinguality}, and suboptimal tokenization \citep{rust-etal-2021-good}. As an alternative for improving monolingual performance with limited native data, we explore leveraging MT models to generate target-language data for monolingual pretraining.

\paragraph{Pretraining on Machine-Translated Data.} Pretraining on MT-derived data has been explored in monolingual settings for Arabic \cite{alcoba-inciarte-etal-2024-utility} and Indic languages \cite{doshi-etal-2024-pretraining}, as well as in multilingual settings \citep{wang2025multilinguallanguagemodelpretraining}, consistently showing downstream performance on par with models pretrained on native text. Most related to our work is \citet{doshi-etal-2024-pretraining}, who pretrained 28M and 85M decoder models and explored CPT of larger LLMs (Gemma-2B, Llama-3-8B) on translationese and native texts, finding MT-derived data competitive with native data. Yet it remains unclear whether MT pretraining benefits larger models and whether CPT on native texts helps when the base model is pretrained on translationese. Our study fills this gap by examining model scaling on MT-derived data (124M–774M), source-side manipulation before translation, and CPT on native texts.

% \section{Target Languages and Machine Translation Models}

% For the source language, we chose English due to its high-resourceness. For target languages, we decided based on several criteria: (1) language is not yet studied in the context of MT pretraining (2) overall data in that language is relatively scarce, (3) availability of open-source MT model, (4) availability of high-quality human-created NLU benchmarks, and (5) presence of a diagnostic benchmark for linguistic knowledge, similar to BLiMP \cite{warstadt-etal-2020-blimp-benchmark}. All are essential for better understanding MT pretraining's generalization potential to native text beyond language modeling performance.

% For MT models, we use OPUS-MT \cite{tiedemann2023democratizing} English $\rightarrow$ Indonesian\footnote{opus-2019-12-18 version accessed at \url{https://huggingface.co/Helsinki-NLP/opus-mt-en-id}} and English $\rightarrow$ Tamil\footnote{opus-2020-07-26 version accessed at \url{https://huggingface.co/Helsinki-NLP/opus-mt-en-dra}}, achieving BLEU score of 38.7 and 4.6 on flores101-devset, respectively \cite{opus}. We use OPUS-MT due to its open-source nature\footnote{CC-BY 4.0}, small model size, and fast inference. 

\begin{figure*}[t]
  \centering
  \includegraphics[width=\textwidth]{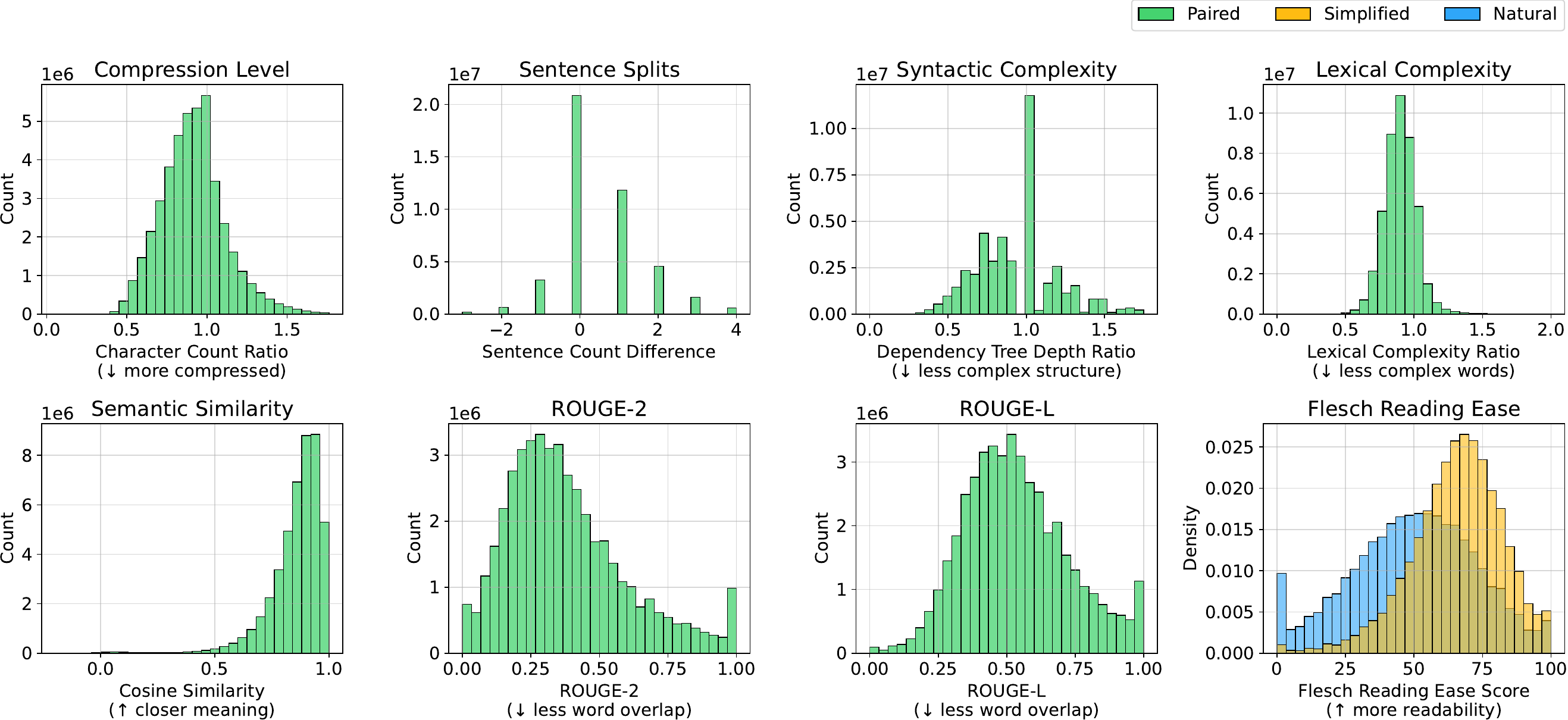}
  \caption{Corpus Feature distributions. Metrics in the first row are adapted from \citet{alva-manchego-etal-2020-asset}. The first row suggests Simplified is shorter, has more sentence splits, uses simpler structures, and uses more common words. The second row shows that Simplified is semantically similar to Natural, with low word-order overlap (low ROUGE-2), moderate preservation of idea flow and structure (moderate ROUGE-L), and clearly higher FRE, indicating systematic differences in readability. For better visualization, we removed outliers, which account for 3\% of the data (see Appendix~\ref{app:outliers} for definition and examples of outliers).}
  \label{fig:corpus-metrics-plots}
\end{figure*}

\section{Data Setup}

\subsection{Languages and MT Systems}
\label{sec:langs-mt}

For the source language, we chose English because of its high-resource status. We selected target languages using the following criteria: (1) the language has not yet been studied in the context of MT pretraining; (2) monolingual data in that language are relatively scarce; (3) an open-source MT model is available; (4) high-quality, human-curated NLU benchmarks exist; and (5) a diagnostic benchmark for linguistic knowledge is available, similar to BLiMP \cite{warstadt-etal-2020-blimp-benchmark}. These criteria are essential for evaluating how MT pretraining generalizes to native text beyond language-modeling performance.

For MT, we use OPUS-MT \cite{tiedemann2023democratizing} for English\,$\rightarrow$\,Indonesian\footnote{Version opus-2019-12-18, \url{https://huggingface.co/Helsinki-NLP/opus-mt-en-id}} and English\,$\rightarrow$\,Tamil\footnote{Version opus-2020-07-26, \url{https://huggingface.co/Helsinki-NLP/opus-mt-en-dra}}, which achieve BLEU scores of 38.7 and 4.6 on the FLORES-101 dev set, respectively \cite{tiedemann-2012-parallel}. We use OPUS-MT due to its open-source license (CC BY 4.0), compact model size, and efficient inference.

% \begin{table}[ht]
% \centering
% \small
% \begin{tabular}{lcccc}
% \toprule
% \textbf{Corpus} & \textbf{Words} & \textbf{Types} & \textbf{TTR} & \textbf{Entropy} \\
% \midrule
% Natural   & 3.72B & 12.70M & 0.34\% & 10.77  \\
% Simplified   & 3.45B & 9.56M & 0.28\% & 10.34 \\
% \bottomrule
% \end{tabular}
% \caption{Source-side corpus statistics. Words are space-separated words, Types are unique word count, TTR is Type-Token Ratio, and Entropy refers to Unigram Entropy. Lower TTR means lower lexical diversity. Lower Entropy means lower complexity.}
% \label{tab:corpus_stats}
% \end{table}

\begin{table}[ht]
\footnotesize
\centering
\begin{tabularx}{\columnwidth}{@{}l S[table-format=2.2] S[table-format=2.2]@{}}

\toprule
\textbf{Feature} & \textbf{Simplified} & \textbf{Natural} \\
\midrule

\multicolumn{3}{@{}l}{\textbf{\textsc{Per-dataset Stats}}} \\
Total words & 3.45B & 3.72B \\
Types (unique words) & 9.56M & 12.70M \\
Type-token ratio (\%)  & 0.28\% & 0.34\% \\
Unigram entropy (bits) & 10.34 & 10.77 \\

\midrule
\multicolumn{3}{@{}l}{\textbf{\textsc{Cross-dataset Stats}}} \\
Compression ($<$80\%) & 27.52\% & \text{\textemdash} \\
Exact match & 2.02\% & \text{\textemdash} \\
High lexical overlap & 3.75\% & \text{\textemdash} \\
Medium lexical overlap & 32.08\% & \text{\textemdash} \\
Low lexical overlap & 60.77\% & \text{\textemdash} \\
Exact mismatch & 1.38\% & \text{\textemdash} \\
Semantic Sim ($>$80\%) & 77.78\% & \text{\textemdash} \\
% Semantic Sim ($>$75\%) & 86.62\% & \text{\textemdash} \\
\bottomrule
\end{tabularx}
\caption{Per-dataset and Cross-dataset statistics of the source-side corpus. Reduced per-dataset stats in Simplified indicate lower complexity compared with Natural. Lexical overlap is measured using ROUGE-2 (R2), with the following thresholds: exact match ($R2 = 1$), high ($0.8 < R2 < 1$), medium ($0.4 < R2 \leq 0.8$), low ($0 < R2 \leq 0.4$), and exact mismatch ($R2 = 0$). Semantic Sim is computed as the cosine similarity of the paragraph embeddings. Cross-dataset stats suggest Simplified texts differ in form but preserve core content. See examples in Appendix \ref{app:examples-semantic} and \ref{app:examples-rouge2}.}
\label{tab:corpus-metrics-table}
\end{table}

\paragraph{Native Data.} For Indonesian, we use Indo4B \cite{wilie-etal-2020-indonlu}, one of the largest and most widely adopted pretraining datasets for the language. For Tamil, we sample 5B tokens from the Tamil subset of IndicMonoDoc \cite{doshi-etal-2024-pretraining}, a large-scale, document-level pretraining corpus.

\paragraph{Natural Data.} The English data was drawn from three permissively licensed corpora\footnote{Dolma and FineWeb-Edu (ODC-BY), Wiki-40B (CC)}: Dolma v1.6 \cite{soldaini-etal-2024-dolma}, FineWeb-Edu \cite{penedo2024finewebdatasetsdecantingweb}, and Wiki-40B \cite{guo-etal-2020-wiki}. The final dataset contains 4B tokens, with 40\% Dolma (web, social media, books, academic), 10\% Wiki-40B (Wikipedia), and 50\% FineWeb-Edu (web).

% \paragraph{Simplified Corpus.} We use Llama 3.1 8B \cite{grattafiori2024llama3herdmodels} to transform Natural Corpus into simplified texts, referred to as Simplified Corpus. We simplify by reducing surface-level complexity—shorter sentences, simpler words, simpler structure—while keeping core text content close to constant. For efficient inference, we use the INT8 quantized version\footnote{ \url{https://huggingface.co/neuralmagic/Meta-Llama-3.1-8B-Instruct-quantized.w8a8}} of the model and vLLM \cite{kwon2023efficientmemorymanagementlarge} as inference server. More details about the prompt and data pipeline in Appendix \ref{}. We confirm it's relative complexity by analyzing per-dataset, cross-dataset metrics (Table~\ref{tab:corpus-metrics-table}), and distributions (Figure~\ref{fig:corpus-metrics-plots}). The following is an example of a simplified text:

\paragraph{Simplified Data.} We use Llama 3.1 8B \cite{grattafiori2024llama3herdmodels} to convert the Natural Data into simplified texts, referred to as the Simplified Data. Simplification reduces surface-level complexity—shorter sentences, simpler words, and simpler structures—while keeping core content approximately constant. For efficient inference, we employ the INT8 quantized version\footnote{\url{https://huggingface.co/neuralmagic/Meta-Llama-3.1-8B-Instruct-quantized.w8a8}} of the model with vLLM \cite{kwon2023efficientmemorymanagementlarge} as the inference server. More details on the prompt in Appendix~\ref{app:prompt}. We validate the reduction in complexity and preservation of core content using per-dataset and cross-dataset metrics (Table~\ref{tab:corpus-metrics-table}) as well as distributional analysis (Figure~\ref{fig:corpus-metrics-plots}). An example simplified text is shown below:

\begin{quote}
    \setlength{\leftskip}{1pt} % Adjust left margin
    \setlength{\rightskip}{1pt} % Adjust right margin
    \setlength{\baselineskip}{0.9\baselineskip} % Reduce line spacing
    \textbf{Natural Data}: Maintaining a relaxed state of mind allows you to approach challenges with clarity and calm, making it easier to find balanced solutions.
    
    \textbf{Simplified Data}: Staying calm helps you face challenges more clearly and find better solutions.
\end{quote}

\paragraph{Machine-Translated Data.} Translation is performed at the sentence level and then reconstructed into documents. We apply pre-MT and post-MT processing and filtering to control quality and efficiency (see Appendix \ref{app:filtering}). Token statistics for all datasets are shown in Table \ref{tab:budgets}.

% Before MT, and apply pre-MT filtering: drop any document that contains a sentence exceeding a given token-count threshold. This is done simply for efficiency purposes. More details in Appendix~\ref{app:filtering}. 

% After MT, we apply post-MT filtering: calculate  sentence length ratio (in tokens) of translation to source text then drop documents in which any translation exceeds a sentence-length ratio of 2. 

% After filtering, sentences are reconstructed back to documents. To control for core text content of the corpus, we ensure that all documents in Natural Corpus and Simplified Corpus are parallel. The final translated corpus will be referred to as Natural-MT Corpus if source data is Natural Corpus and Simplified-MT Corpus if source data is Simplified Corpus.

\subsection{Evaluation and Fine-tuning Data}
The evaluation touches on three aspects: 
(1) out-of-distribution generalization to native text, 
(2) native-language proficiency, and 
(3) native-language downstream performance.

\textbf{Aspect (1): Out-of-distribution generalization to native text.}  
We use a held-out validation set of 200 million tokens from each language's native corpus and compute cross-entropy loss. Strong performance indicates proficiency in native language modeling.

\textbf{Aspect (2): Native-language grammatical proficiency.}  
We use the LINDSEA syntax subset \cite{leong2023bhasaholisticsoutheastasian}, formatted as minimal pairs—sentence pairs differing only by a specific grammatical feature to test whether a model favors the grammatical form over the ungrammatical one. The benchmark covers morphology, negation, argument structure, and filler-gap dependencies. Strong performance indicates robust grammatical knowledge.

\textbf{Aspect (3): Native-language NLU performance.}  
We evaluate on the Indonesian and Tamil subsets of SEA-HELM \cite{susanto2025seahelmsoutheastasianholistic} across four NLU tasks: sentiment analysis (SA), toxicity detection (TD), natural language inference (NLI), and causal reasoning (CR). Strong performance indicates effective transfer from MT-derived to native data.

\subsection{Fine-tuning Data}

% \begin{table}[ht]
% \centering
% \small
% \begin{tabular}{lcccc}
% \toprule
% \textbf{Task} & \textbf{Train Data} & \textbf{Labels} & \textbf{Total} \\
% \midrule
% Natural   & 3.72B & 12.70M & 0.34\%   \\
% Simplified   & 3.45B & 9.56M & 0.28\%  \\
% \bottomrule
% \end{tabular}
% \caption{Source-side corpus statistics. Words are space-separated words, Types are unique word count, TTR is Type-Token Ratio, and Entropy refers to Unigram Entropy. Lower TTR means lower lexical diversity. Lower Entropy means lower complexity.}
% \label{tab:downstream_datasets}
% \end{table}

\begin{table}[ht]
\centering
\small
\setlength{\tabcolsep}{6pt}        % horizontal padding
\renewcommand{\arraystretch}{1.3}  % vertical padding

\begin{tabularx}{\columnwidth}{@{}l X X c@{}}
\toprule
\textbf{Task} & \textbf{Train Data} & \textbf{Labels (counts)} \\
\midrule

\textit{SA}
  & \makecell[l]{Amazon\\\cite{hou2024bridginglanguageitemsretrieval} \\Yelp\\\cite{NIPS2015_250cf8b5}}
  & \makecell[l]{negative (50K)\\positive (50K)} \\
\addlinespace

\textit{TD}
  & \makecell[l]{HateSpeech\\\cite{davidson2017automatedhatespeechdetection}}
  & \makecell[l]{hate (0.6K)\\clean (2.4K)\\rough (10.3K)} \\
\addlinespace

\textit{NLI}
  & \makecell[l]{WANLI\\\cite{liu2022wanliworkeraicollaboration}}
  & \makecell[l]{contradiction (11.2K)\\entailment (10.9K)\\neutral (11K)} \\
\addlinespace
\textit{CR}
  & \makecell[l]{B-COPA\\\cite{kavumba2019choosingplausiblealternativesclever}}
  & \makecell[l]{cause (0.5K)\\effect (0.5K)} \\

\bottomrule
\end{tabularx}

\caption{Overview of fine-tuning tasks, data sources, label splits, and example counts (in thousands). SA = Sentiment Analysis, TD = Toxicity Detection, NLI = Natural Language Inference, CR = Causal Reasoning.}
\label{tab:finetune-tasks}
\end{table}

In low-resource settings with little or no fine-tuning data, we extend the MT pretraining approach by translating English task datasets into the target language (Table~\ref{tab:finetune-tasks}). All datasets are curated to be label-balanced, except TD, where downsampling would reduce the data to roughly 600 examples per label. Translation and filtering follow the same procedure as used for pretraining data.

% To investigate the generalization potential of machine-translated fine-tuning data, we train on machine-translated English task-specific datasets. The list of datasets is summarized in Table~\ref{tab:finetune-tasks}. All task datasets, except for causal reasoning, will go through balanced-label sampling $\rightarrow$ pre-MT filtering $\rightarrow$ MT $\rightarrow$ post-MT filtering. For more details, refer to Appendix~\ref{app:filtering}.

\section{Experimental Setup}

\subsection{Models and Training}
\label{sec:models-training}

\paragraph{Architectures.}
We train models in three sizes (Table~\ref{tab:arch}) following the GPT-2 architecture \cite{radford2019language}. A 50{,}257-token BPE \cite{sennrich2016neuralmachinetranslationrare} is trained per language on native data and reused across all pretraining conditions (Native, Natural-MT, Simplified-MT). Details on the tokenizer and special tokens are provided in Appendix~\ref{app:train-details}.

\begin{table}[ht]
\centering
\small
\begin{tabular}{lrrrrr}
\toprule
\textbf{Size} & \textbf{Layers} & \textbf{$d_{\text{model}}$} & \textbf{Heads} &
\textbf{MLP} & \textbf{Params} \\
\midrule
Small  & 12 & 768  & 12 & 3072 & 124M \\
Medium & 24 & 1024 & 16 & 4096 & 355M \\
Large  & 36 & 1280 & 20 & 5120 & 774M \\
\bottomrule
\end{tabular}
\caption{Model configurations for the three GPT-2 sizes. Columns show number of layers, hidden size ($d_{\text{model}}$), attention heads, feed-forward dimension (MLP), and parameter counts in millions.}
\label{tab:arch}
\end{table}

\paragraph{Pretraining conditions.}
For each language we train nine models: three corpora (Native, Natural-MT, Simplified-MT) crossed with three sizes (Small, Medium, Large). We use causal language modeling objective with a 1{,}024-token context. Native-only models are pretrained on whole native corpus (4.3B for Indonesian and 5B for Tamil) to serve as a proxy for upper bound performance in low-resource scenarios. Full optimizer and schedule details are in Appendix~\ref{app:train-details}.

\paragraph{Continual pretraining (CPT).}
For CPT, we continue pretraining the final Natural-MT and Simplified-MT models on a subset of native corpus (1B tokens for Indonesian, 2.5B for Tamil). All settings match pretraining except for a lower peak learning rate. More details in Appendix~\ref{app:train-details}.

% For Natural-MT and Simplified-MT checkpoints, we continue training on native text from the final MT checkpoint (1B native tokens for Indonesian; 2.5B for Tamil). All settings mirror pretraining except for a reduced peak learning rate; see Appendix~\ref{app:train-details}.

\paragraph{Token budgets.}
Table~\ref{tab:budgets} summarizes MT and native token budgets for each training setup. CPT refers to native continuation after MT pretraining stage. For example, in Indonesian, Native-only is trained on 4.3B native tokens, Natural-MT on 2.9B MT-derived tokens, and Natural-MT-CPT continues Natural-MT training with an additional 1B native tokens.

% Table~\ref{tab:budgets} summarizes MT and native token budgets per setup. CPT denotes the native continuation after an MT pretraining stage.

% \begin{table}[ht]
% \centering
% \small
% \setlength{\tabcolsep}{3pt}
% \renewcommand{\arraystretch}{1.2}
% \begin{tabularx}{\columnwidth}{@{}l
%   >{\centering\arraybackslash}X >{\centering\arraybackslash}X
%   >{\centering\arraybackslash}X >{\centering\arraybackslash}X@{}}
% \toprule
% & \multicolumn{2}{c}{\textbf{Indonesian}} & \multicolumn{2}{c}{\textbf{Tamil}} \\
% \cmidrule(lr){2-3}\cmidrule(lr){4-5}
% \textbf{Setup} & {MT} & {Native} & {MT} & {Native} \\
\begin{table}[t]
\centering
\footnotesize
\setlength{\tabcolsep}{3pt}
\begin{tabularx}{\linewidth}{@{\extracolsep{\fill}}
  l
  c
  c
  c
  c
@{}}
\toprule
& \multicolumn{2}{c}{\textbf{Indonesian}} & \multicolumn{2}{c}{\textbf{Tamil}} \\
\cmidrule(lr){2-3}\cmidrule(lr){4-5}
\textbf{Setup} & \multicolumn{1}{c}{MT} & \multicolumn{1}{c}{Native} &
                 \multicolumn{1}{c}{MT} & \multicolumn{1}{c}{Native} \\
\midrule
Native          & \text{\textemdash} & 4.3B & \text{\textemdash} & 5.0B \\
\pairsplit
Natural-MT      & 2.9B & \text{\textemdash} & 4.8B & \text{\textemdash} \\
Natural-MT-CPT    & 2.9B & 1.0B & 4.8B & 2.5B \\
\pairsplit
Simplified-MT   & 2.7B & \text{\textemdash} & 5.2B & \text{\textemdash} \\
Simplified-MT-CPT & 2.7B & 1.0B & 5.2B & 2.5B \\
\bottomrule
\end{tabularx}
\caption{Training token budgets by setup for each language (billions). MT counts reflect machine-translated corpora; Native counts reflect native-language text. CPT denotes native continuation from the MT checkpoint. All token counts are computed with each language's fixed 50{,}257-token BPE tokenizer trained on native corpora and reused across all conditions.}
\label{tab:budgets}
\end{table}

\subsection{Fine-tuning \& Evaluation}
\paragraph{Supervised tasks.}
Each pretrained checkpoint is fine-tuned on \textit{sentiment analysis} (SA), \textit{natural-language inference} (NLI), and \textit{toxicity detection} (TD; Indonesian only) using machine-translated training data, then evaluated on native SEA-HELM test sets. Dataset sources and label splits are in Table~\ref{tab:finetune-tasks}. We also fine-tune on \textit{causal reasoning} (CR), but because all systems remain near chance ($\approx$50–54\% balanced accuracy) with no clear trends, we omit CR from the main results tables; for transparency, full CR means\,$\pm$\,std appear in Appendix Table~\ref{tab:finetune_std}.

\paragraph{No pretraining baseline.}
For each size (Small/Medium/Large), we also train a \textit{No Pretraining} baseline: a randomly initialized GPT-2 decoder with the same architecture and classification head, optimized only on the task data (no LM pretraining). Optimization settings, sequence length, and hyperparameter search match those used for pretrained checkpoints.

\paragraph{Metric and model selection.}
We select by \textbf{balanced accuracy} on a translationese dev split and report average scores over three seeds on SEA-HELM benchmark. Batch sizes per task are listed in Appendix Table~\ref{tab:batchsizes}; fine-tuning heads, pooling, and the hyperparameter search space are described in Appendix~\ref{app:finetune-settings}.

\paragraph{Zero-shot syntactic probing.}
To assess the linguistic knowledge encoded in the pretrained representations, we evaluate all models on the Syntax subset of LINDSEA.  The subset is converted to BLiMP-style minimal pairs; a model is correct when it assigns a higher log-probability to the grammatical member of the pair. Accuracy is averaged across all syntactic phenomena.

\begin{table}[t]
\centering
\footnotesize
\setlength{\tabcolsep}{3pt}
\begin{tabular*}{\linewidth}{@{\extracolsep{\fill}}
  l
  S[table-format=2.1]
  S[table-format=+2.1]
  S[table-format=2.1]
  S[table-format=+2.1]
@{}}
\toprule
& \multicolumn{2}{c}{\textbf{Indonesian}} & \multicolumn{2}{c}{\textbf{Tamil}} \\
\cmidrule(lr){2-3}\cmidrule(lr){4-5}
\textbf{Model} & \multicolumn{1}{c}{Acc.} & \multicolumn{1}{c}{$\Delta$} &
                 \multicolumn{1}{c}{Acc.} & \multicolumn{1}{c}{$\Delta$} \\
\midrule
\multicolumn{5}{l}{\textbf{Small}}\\
\pairsplit
\hspace{0.75em}Native                    & {\bfseries 53.6} & {}    & 71.5 & {} \\
\pairsplit
\hspace{0.75em}Natural-MT                & 47.6             & {}    & 66.2 & {} \\
\hspace{0.75em}Natural-MT-CPT    & 52.9             & \gain{+5.3}  & 69.1 & \gain{+2.9} \\
\pairsplit
\hspace{0.75em}Simplified-MT             & 46.6             & {}    & 61.3 & {} \\
\hspace{0.75em}Simplified-MT-CPT & 52.4             & \gain{+5.8}  & {\bfseries 72.1} & \gain{+10.8} \\
\midrule
\multicolumn{5}{l}{\textbf{Medium}}\\
\pairsplit
\hspace{0.75em}Native                    & 52.4             & {}    & 62.8 & {} \\
\pairsplit
\hspace{0.75em}Natural-MT                & 50.5             & {}    & 65.5 & {} \\
\hspace{0.75em}Natural-MT-CPT    & {\bfseries 53.7} & \gain{+3.2}  & 72.8 & \gain{+7.3} \\
\pairsplit
\hspace{0.75em}Simplified-MT             & 49.5             & {}    & 65.1 & {} \\
\hspace{0.75em}Simplified-MT-CPT & 52.1             & \gain{+2.6}  & {\bfseries 76.0} & \gain{+10.9} \\
\midrule
\multicolumn{5}{l}{\textbf{Large}}\\
\pairsplit
\hspace{0.75em}Native                    & {\bfseries 57.4} & {}    & 70.9 & {} \\
\pairsplit
\hspace{0.75em}Natural-MT                & 49.7             & {}    & 62.8 & {} \\
\hspace{0.75em}Natural-MT-CPT    & 54.5             & \gain{+4.8}  & {\bfseries 72.8} & \gain{+10.0} \\
\pairsplit
\hspace{0.75em}Simplified-MT             & 49.7             & {}    & 62.8 & {} \\
\hspace{0.75em}Simplified-MT-CPT & 56.3             & \gain{+6.6}  & 70.9 & \gain{+8.1} \\
\bottomrule
\end{tabular*}
\caption{%
Accuracy on the LINDSEA Syntax subset (higher is better; random chance is 50\,\%). Native pretraining produces the strongest Indonesian model (57.4\%), whereas CPT lifts MT models to the top for Tamil (76.0\% for Medium Simplified-MT-CPT). In Indonesian, MT models score close to or below random, but CPT raises them by 2–7 percentage points, partially closing the gap to native. Tamil results are uniformly higher: even MT-only models exceed 60\%, and CPT adds another 7–11 percentage points. Medium Simplified-MT-CPT surpasses all Large models in Tamil. A per-phenomenon breakdown appears in Appendix Table~\ref{tab:syntax_breakdown}.
}
\label{tab:syntax}
\end{table}

\section{Results and Discussion}

We present results by our three research questions, then report translationese fine-tuning outcomes. Each subsection starts with a short answer, followed by evidence and a practical takeaway.

% \subsection{RQ1 — Does scaling on MT reduce loss on native text?}
\subsection{Does scaling on MT-derived data improve loss on native text?}
\label{sec:rq1}

\noindent\textbf{Answer:} Within our setup, yes. Larger MT-pretrained models generally achieve lower loss on held-out native text than smaller ones, except for the Tamil Simplified-MT 774M model, which performs slightly worse. 

\noindent\textbf{Evidence:} For both languages, validation loss on native text decreases with larger model size when pretrained on MT-derived data (Fig.~\ref{fig:cpt}). Diminishing returns appear at 774M, likely due to the data–to–parameter ratio, but further experiments are needed to confirm. Overall, the trend suggests larger models improve generalization to native text, despite being trained only on MT-derived data. This pattern persists after CPT, indicating that greater capacity captures transferable structure rather than simply memorizing translation artifacts.

\noindent\textbf{Takeaway:} More parameters enhance transfer to native text even when pretraining solely on MT-derived data.

\begin{table}[ht]
\centering
\footnotesize
\setlength{\tabcolsep}{3pt}
\begin{threeparttable}
\begin{tabular*}{\linewidth}{@{\extracolsep{\fill}}
  l
  S[table-format=2.1] % ID Sentiment
  S[table-format=2.1] % ID NLI
  S[table-format=2.1] % ID Toxicity
  S[table-format=2.1] % TA Sentiment
  S[table-format=2.1] % TA NLI
@{}}
\toprule
& \multicolumn{3}{c}{\textbf{Indonesian}} & \multicolumn{2}{c}{\textbf{Tamil}} \\
\cmidrule(lr){2-4}\cmidrule(lr){5-6}
\textbf{Model} &
\multicolumn{1}{c}{SA} &
\multicolumn{1}{c}{NLI} &
\multicolumn{1}{c}{TD} &
\multicolumn{1}{c}{SA} &
\multicolumn{1}{c}{NLI} \\
\midrule
\multicolumn{6}{l}{\textbf{Small}}\\
\pairsplit
\hspace{0.75em}No Pretraining (LB)                        & 56.1 & 43.0 & 41.3 & 75.3 & 38.3 \\
\hspace{0.75em}Native (UB)                                & 63.4 & 53.7 & {\bfseries 52.6} & 87.1 & 42.8 \\
\pairsplit
\hspace{0.75em}Natural-MT                            & 61.9 & 56.9 & 42.5 & 88.4 & 42.3 \\
\hspace{0.75em}Natural-MT-CPT           & {\bfseries 63.5} & 57.4 & 47.6 & 88.9 & {\bfseries 43.5} \\
\pairsplit
\hspace{0.75em}Simplified-MT                         & 61.3 & 56.2 & 44.5 & 88.8 & 40.7 \\
\hspace{0.75em}Simplified-MT-CPT        & 62.9 & {\bfseries 58.2} & 49.6 & {\bfseries 89.0} & 43.0 \\
\midrule
\multicolumn{6}{l}{\textbf{Medium}}\\
\pairsplit
\hspace{0.75em}No Pretraining (LB)                       & 55.9 & 43.7 & 41.8 & 75.2 & 38.9 \\
\hspace{0.75em}Native (UB)                                & 62.7 & 57.7 & {\bfseries 53.0} & 84.8 & 41.1 \\
\pairsplit
\hspace{0.75em}Natural-MT                            & 62.6 & {\bfseries 60.7} & 44.1 & 90.3 & 43.8 \\
\hspace{0.75em}Natural-MT-CPT           & {\bfseries 64.2} & 59.7 & 49.5 & {\bfseries 91.2} & {\bfseries 45.1} \\
\pairsplit
\hspace{0.75em}Simplified-MT                         & 61.6 & 55.8 & 44.6 & 90.6 & 44.8 \\
\hspace{0.75em}Simplified-MT-CPT        & 62.6 & 57.2 & 48.3 & 90.5 & {\bfseries 45.1} \\
\midrule
\multicolumn{6}{l}{\textbf{Large}}\\
\pairsplit
\hspace{0.75em}No Pretraining (LB)                        & 56.0 & 37.1 & 41.0 & 75.8 & 40.0 \\
\hspace{0.75em}Native (UB)                                & 63.7 & 56.6 & {\bfseries 54.7} & 86.2 & 43.4 \\
\pairsplit
\hspace{0.75em}Natural-MT                            & 62.6 & 61.6 & 45.2 & 90.6 & 43.6 \\
\hspace{0.75em}Natural-MT-CPT           & 63.7 & 61.4 & 48.3 & {\bfseries 92.1} & {\bfseries 45.6} \\
\pairsplit
\hspace{0.75em}Simplified-MT                         & 61.5 & {\bfseries 63.2} & 46.2 & 90.0 & 43.3 \\
\hspace{0.75em}Simplified-MT-CPT        & {\bfseries 64.3} & 61.9 & 49.1 & 90.3 & 44.4 \\
\bottomrule
\end{tabular*}
  % \caption[Balanced accuracy means]{%
  %   Average balanced accuracy on the SEA-HELM test sets after fine-tuning each model on translationese over three random seeds (best configuration per seed). For \textbf{sentiment analysis (SA)} and \textbf{natural language inference (NLI)}, the gap between MT-pretrained models and native-pretrained models is narrow; continual pretraining (CPT) usually nudges performance to the top of each size block. \textbf{Toxicity detection (TD)} clearly favours native-pretraining; MT-pretrained models lag by 3–11 percentage points, despite identical fine-tuning data.}
  \caption[Balanced accuracy means]{%
    Balanced accuracy on SEA-HELM after fine-tuning each model on translationese (averaged over three seeds). \textbf{LB} = lower bound (No Pretraining); \textbf{UB} = upper bound (Native). For \textbf{SA} and \textbf{NLI}, MT-pretrained models approach Native performance, with CPT typically boosting results beyond UB. For \textbf{TD}, Native pretraining remains stronger, with MT-pretrained models lagging by 3–11 points despite identical fine-tuning data. Standard deviations are in Table~\ref{tab:finetune_std} in the Appendix.}
    \label{tab:finetune_results}
    % \begin{tablenotes}
    %   \footnotesize
    %   \item Standard deviations over the three random seeds are reported in Table~\ref{tab:finetune_std} in the appendix.
    % \end{tablenotes}
  \end{threeparttable}
\end{table}

\subsection{Does source-side simplification help transfer to native text?}
% \subsection{RQ2 — Does source-side simplification help transfer?}
\label{sec:rq2}

\noindent\textbf{Answer:} Within our setup, no. Simplifying English before translation reduces transfer to native text.

\noindent\textbf{Evidence:} In language modeling, Simplified-MT yields worse loss on native text than Natural-MT across all sizes (see Fig.~\ref{fig:cpt}). In syntactic probing, Natural-MT consistently outperforms Simplified-MT, with the largest gap in Tamil small models, though the gap narrows with larger sizes (Table~\ref{tab:syntax}). In downstream tasks, neither is consistently better—Simplified-MT leads on some tasks and Natural-MT on others—except for TD, which strongly favors Native models. Overall, accuracy differences are usually within 1–2 points (Table~\ref{tab:finetune_results}), suggesting that improvements in language modeling loss do not always translate directly into downstream gains.

\noindent\textbf{Takeaway:} For source-side English, higher lexical and syntactic diversity yields MT-derived data that transfers better to native text. Avoid operations that reduce this diversity (e.g., simplification) if the goal is native transfer.

\subsection[MT Pretrain to Native CPT]{Is MT pretrain $\rightarrow$ Native CPT more data-efficient than native-only?}
% \subsection[MT Pretrain to Native CPT]{RQ3 — Is MT pretrain $\rightarrow$ native CPT more data-efficient than native-only?}
\label{sec:rq3}

\noindent\textbf{Answer:} Within our setup, yes. With the same native-token budget, MT-initialized CPT matches or surpasses native-only.

\noindent\textbf{Evidence:} A short CPT phase (1B tokens for Indonesian; 2.5B for Tamil) reduces loss on native text, surpassing native-only models trained on the same native budget. Notably, Tamil CPT models surpassed native-only models trained on 5B native tokens (see Figure ~\ref{fig:cpt}). In syntactic probing, CPT yields significant gains across model sizes, raising accuracy by about 2–7 points in Indonesian and 7–11 points in Tamil (Table~\ref{tab:syntax}). We surmise the gains come from better alignment with the native distribution, suggesting an "error correction" or unlearning of translationese artifacts.

\noindent\textbf{Takeaway:} When native data is scarce, MT pretraining followed by continual pretraining on native text often outperforms native-only pretraining.

\subsection{Translationese fine-tuning outcomes}
\label{sec:rq4}

% \noindent\textbf{Answer:} Accuracy is near parity with Native models on sentiment analysis and natural language inference, while toxicity detection strongly favors Native models.

\noindent\textbf{Answer:} For SA and NLI, MT-pretrained models approach the Native upper bound, with CPT often pushing results beyond it. For TD, performance strongly favors Native models.

\noindent\textbf{Evidence:}
After fine-tuning on translationese, all pretrained models (\textit{Native}, \textit{MT}, \textit{MT-CPT}) exceed the \textit{No Pretraining} baseline across tasks, confirming the utility of pretraining. For \textit{SA} and \textit{NLI}, MT-pretrained models are typically within 1–2 points above the Native models, and CPT variants often \textit{exceed} the upper bound performance (Native) within each size group (Table~\ref{tab:finetune_results}). For Indonesian \textit{TD}, Native models retain a 3–11 point edge over MT-pretrained ones despite identical fine-tuning data. We omit \textit{CR} from Table~\ref{tab:finetune_results} because all systems remain near chance ($\approx$50–54\% balanced accuracy) and perform similarly to \textit{No Pretraining}; full means$\pm$std over three seeds appear in Appendix Table~\ref{tab:finetune_std}.

\noindent\textbf{Takeaway:} In low-resource scenarios, MT-derived fine-tuning data is useful for tasks like sentiment analysis and NLI but has limited value for more culturally nuanced tasks such as toxicity detection.

\section{Conclusion}

In this work, we asked whether larger models improve generalization to native text when pretraining data is pure machine-translated text, how source-side complexity affects transfer to native text, and whether MT-pretrained models are good starting points for continually pretraining on native text. We observed three consistent patterns. First, for the 124M to 774M parameters setup, more parameters improve transfer to native text even when pretraining solely on MT-derived data. Second, for source-side English texts, higher lexical and syntactic diversity yields MT-derived data that transfers better to native text. Avoid operations that reduce this diversity (e.g., simplification) if the goal is native transfer. Third, when native data is scarce, MT pretraining followed by continual pretraining on native text often outperforms native-only pretraining. In scenarios with zero or limited fine-tuning data, MT-derived fine-tuning data is useful for tasks like sentiment analysis and NLI but has limited value for more culturally nuanced tasks such as toxicity detection.

We distill our findings into a recipe for improving monolingual models beyond what is achievable with the available native data:
\begin{itemize}
  \item Generate more target-language data via MT.
  \item Pretrain on MT-derived data (using the largest model size you can afford).
  \item Continue pretraining on native data from an MT-pretrained checkpoint. 
  \item With limited native fine-tuning data and a fixed annotation budget, maximize coverage by translating training data from high-resource languages for tasks like sentiment analysis and NLI, while reserving native annotation for more culturally nuanced tasks like toxicity detection.
\end{itemize}

For future work, extending these experiments to larger models, better MT systems, different source-side and target languages, and more advanced preprocessing that balances MT ease with linguistic diversity will clarify when the effects observed here amplify or taper. Furthermore, extending this approach to post-training regimes such as instruction tuning and preference alignment remains an open direction.

\section*{Limitations}
Our study has some limitations. First, we used a fixed dataset and only three GPT-2 sizes (124M, 355M, 774M), which may limit generalizability; broader variation in data and scale could yield different insights. Second, fine-tuning relied on translated rather than native data, so it is unclear if the same patterns hold with native training data. Third, MT quality matters—BLEU scores varied across languages, but we did not separate translation effects from linguistic confounds. Fourth, LLM-based simplification can hallucinate or omit information, causing Simplified-MT to diverge semantically from Natural-MT to some degree. Finally, since language and culture are deeply connected, our focus on translation does not address the transfer of cultural knowledge.

%Seventh, no qualitative analysis on generation quality. Eight, alignment is not explored and the question remains whether the learned representations will help in alignment.

\bibliography{anthology,custom}
%\bibliography{custom}

\appendix

\section{Examples from Natural and Simplified Data by Semantic Similarity}
\label{app:examples-semantic}
As shown in Table \ref{tab:corpus-metrics-table}, 77.78\% of datasets have semantic similarity of greater than 80\%. We show examples here of texts with varying semantic similarity scores with their corresponding ROUGE-2 scores.

\noindent
Examples of semantic similarity > 0.8:
\lstset{
  basicstyle=\ttfamily\small,
  breaklines=true,
  columns=fullflexible,
  frame=single,
  backgroundcolor=\color{gray!10},
}
\begin{lstlisting}
SEMANTIC SIMILARITY: 0.90, ROUGE-2: 0.27;
	Natural:important officials and well known persons who visited the islands wrote
	Simplified:important visitors to the islands wrote
SEMANTIC SIMILARITY: 0.95, ROUGE-2: 0.41;
	Natural:Also, the authors now expect to apply their approach to other regions. They have a lot of work to do. After all, arid landscapes occupy about 65 million square kilometers of the earth's surface (this is almost four areas of Russia).
	Simplified:The authors now plan to use their method in other areas. They have a lot of work ahead of them. Arid landscapes cover almost 65 million square kilometers of the Earth's surface, which is roughly four times the size of Russia.
SEMANTIC SIMILARITY: 0.90, ROUGE-2: 0.19;
	Natural:On its face, the USDA's decision to have participation in the NAIS be voluntary seems to solve all of the major concerns. Small and organic farmers will be able to "opt out" of participation in the NAIS if they have objections to its methodology. [FN203]
	Simplified:The USDA made the NAIS voluntary. This means that small and organic farmers can choose not to participate if they don't agree with how the NAIS works.
SEMANTIC SIMILARITY: 0.96, ROUGE-2: 0.43;
	Natural:The ICD-11 includes a revised definition for alcohol use disorders (AUDs) and, more specifically, for alcohol dependence and the "harmful patterns of alcohol use."
	Simplified:The ICD-11 has changed how it defines alcohol use disorders (AUDs). It now includes a new definition for alcohol dependence and for when alcohol use causes harm.
SEMANTIC SIMILARITY: 0.95, ROUGE-2: 0.75;
	Natural:Feel free to check out more of this website. Our goal is to provide rebuttals to the bad science behind young earth creationism, and honor God by properly presenting His creation.
	Simplified:Our goal is to provide rebuttals to the bad science behind young earth creationism, and honor God by properly presenting His creation. You can find more information on this website.
SEMANTIC SIMILARITY: 0.82, ROUGE-2: 0.50;
	Natural:separate trees you simply set the CODEBASE attributes of each applet
	Simplified:set the CODEBASE attribute of each applet
SEMANTIC SIMILARITY: 0.98, ROUGE-2: 0.74;
	Natural:The U.S. Geological Survey's National Wildlife Health Center verified the disease in a little brown bat found this month in North Bend, about 30 miles east of Seattle.
	Simplified:The U.S. Geological Survey's National Wildlife Health Center found a disease in a little brown bat in North Bend, which is about 30 miles east of Seattle.
\end{lstlisting}

\noindent
Examples of semantic similarity < 0.5:
\lstset{
  basicstyle=\ttfamily\small,
  breaklines=true,
  columns=fullflexible,
  frame=single,
  backgroundcolor=\color{gray!10},
}
\begin{lstlisting}
SEMANTIC SIMILARITY: 0.09, ROUGE-2: 0.00;
	Natural:- Press Ctrl + 2 to add more text boxes. Press Ctrl + shift + 2 to adjust text box.
	Simplified:(Note: Please provide your output in the format specified above, ensuring it is free of grammatical errors and easy to read.)
SEMANTIC SIMILARITY: 0.38, ROUGE-2: 0.00;
	Natural:his bark is worse than his bite, he is bad-tempered but harmless
	Simplified:This person is grumpy, but he won't hurt you.
SEMANTIC SIMILARITY: 0.44, ROUGE-2: 0.00;
	Natural:said to have sworn, under duress, that he
	Simplified:The person was forced to say something, but he didn't really mean it.
SEMANTIC SIMILARITY: 0.35, ROUGE-2: 0.24;
	Natural:and operated at 33 MHz and 20 MIPS. ...Many thanks to Robert B Garner - who
	Simplified:The computer was made by Intel and operated at 33 million cycles per second and 20 million instructions per second.
SEMANTIC SIMILARITY: 0.48, ROUGE-2: 0.32;
	Natural:you are near the surface of the Earth, regardless of what the object is
	Simplified:The surface of the Earth is the outermost solid layer of our planet.
SEMANTIC SIMILARITY: 0.36, ROUGE-2: 0.09;
	Natural:upon his visage, rather than pure devotion, such as one might
	Simplified:The person's face showed more of a sense of duty than pure love.
SEMANTIC SIMILARITY: 0.14, ROUGE-2: 0.00;
	Natural:- Genetic screens in human cells using the CRISPR-Cas9 system. Science 343, 80-84 (2014) , , &
	Simplified:Simplification of the text should be provided in the format specified above.
SEMANTIC SIMILARITY: 0.11, ROUGE-2: 0.00;
	Natural:Strategies you implement are usually defined as the tone of your information. Here is the summary of tone types:
	Simplified:(Note: Please provide your output in the format specified above, ensuring it is clear, well-organized, and free of grammatical errors.)
SEMANTIC SIMILARITY: 0.08, ROUGE-2: 0.00;
	Natural:- Mathematics - Knowledge of arithmetic, algebra, geometry, calculus, statistics, and their applications.
	Simplified:Simplification of the text should be done in the same format as the examples provided.
SEMANTIC SIMILARITY: 0.14, ROUGE-2: 0.00;
	Natural:Art. 304, consists of two clauses, and each clause operates as a proviso to Arts. 301 and 303.
	Simplified:The law has two parts. Each part is connected to other laws.
SEMANTIC SIMILARITY: 0.45, ROUGE-2: 0.00;
	Natural:- Can you think of other cases where a government has addressed its previous wrongdoing?
	Simplified:- Yes, there are several examples.
\end{lstlisting}

\section{Examples from Natural and Simplified Data by ROUGE-2}
\label{app:examples-rouge2}
In Table \ref{tab:corpus-metrics-table}, we used ROUGE-2 (R2) thresholds to define the level of lexical overlap. 

\noindent
\textbf{Examples of low lexical overlap ($0 < R2 \leq 0.4$):}

\lstset{
  basicstyle=\ttfamily\small,
  breaklines=true,
  columns=fullflexible,
  frame=single,
  backgroundcolor=\color{gray!10},
}
\begin{lstlisting}
ROUGE-2: 0.19;
	Natural:An independent panel of technical experts convened by the American Chemical Society Green Chemistry Institute formally judged the 2017 submissions from among scores of nominated technologies and made recommendations to EPA for the 2017 winners. The 2017 awards event will be held in conjunction with the 21st Annual Green Chemistry and Engineering Conference.
	Simplified:An independent group of experts looked at many technologies and chose the best ones for the 2017 awards. They recommended these winners to the EPA. The 2017 awards ceremony will be held at the same time as a conference on green chemistry.
ROUGE-2: 0.38;
	Natural:Only $24.00 and a pair of high boots was all it took for the first property owner to purchase the land where the now renowned Pioneer Courthouse Square is located. The block was the site for Portland's first school. Shortly thereafter, it became the Portland Hotel where it served as a social center. The hotel was demolished in 1951 to make room for the automobile with installation of a full city block of parking. Due to progressive civic leadership in the 1970's, Portland worked to revitalize its downtown, including a move away from the use of automobiles and back toward mass transit. The demolition of the parking garage and creation of Pioneer Courthouse Square remains a major landmark of this effort.
	Simplified:Only $24.00 and a pair of boots was all it took for the first person to buy the land where Pioneer Courthouse Square is now. This block was once home to Portland's first school. Later, it became the Portland Hotel, where people would meet and socialize. The hotel was torn down in 1951 to make room for cars. In the 1970s, Portland's leaders decided to make the city more people-friendly. They wanted to reduce the use of cars and increase the use of public transportation. As part of this effort, the parking garage was removed, and Pioneer Courthouse Square was created.
ROUGE-2: 0.10;
	Natural:- 2002 - 2011 is the ten years preceding the ratings evaluation, and
	Simplified:- 2002 to 2011 was the time before the ratings were checked.
ROUGE-2: 0.39;
	Natural:The wearing of gowns at formals is compulsory at some colleges and various other traditions are usually observed, including grace said in Latin or English. The wearing of gowns may sometimes constitute the only dress code; in other cases, formal wear (for example, a lounge suit for men or equivalent for women) is required in addition to, or instead of, the gown.
	Simplified:The wearing of gowns at formals is required at some colleges and some other traditions are followed, like saying grace in Latin or English. In some places, wearing a gown is the only dress code, while in others, you also need to wear formal clothes (like a suit for men or something similar for women) along with the gown.
\end{lstlisting}

\noindent
\textbf{Examples of medium lexical overlap  ($0.4 < R2 \leq 0.8$)}:
\lstset{
  basicstyle=\ttfamily\small,
  breaklines=true,
  columns=fullflexible,
  frame=single,
  backgroundcolor=\color{gray!10},
}
\begin{lstlisting}
ROUGE-2: 0.68;
	Natural:HDTV technology is estimated that this will be the future of television standards, so a senior researcher in the field of systems and management strategies Dr. Indu Singh predicts that the world market for HDTV would reach 250 billion dollars per year (year 2010).
	Simplified:HDTV technology is expected to be the future of television standards. Dr. Indu Singh, a senior researcher in the field of systems and management strategies, predicts that the world market for HDTV will reach $250 billion per year by 2010.
ROUGE-2: 0.74;
	Natural:Prophetically, he feels the need to plead for ten years of life so that:
	Simplified:Prophetically, he feels the need to ask for ten more years of life so that:
ROUGE-2: 0.47;
	Natural:Most common palm species are Elaeis guineensis and Borassus aethiopium (rhun palm).
	Simplified:The two most common types of palm trees are Elaeis guineensis and Borassus aethiopium, also known as the rhun palm.
ROUGE-2: 0.51;
	Natural:The glare of publicity that swirled about Yellow Thunder Camp last September when the government ordered its occupants to leave their chosen spot has faded like the leaves of autumn. The traditional but transient tepees have been supplemented with a geodesic dome. The legal battle which will determine the camp's future drags on in nearby Rapid City.
	Simplified:The glare of publicity that swirled around Yellow Thunder Camp last September when the government ordered its occupants to leave their chosen spot has faded. The campers have added a new, dome-shaped shelter to their traditional tepees. The legal fight about the camp's future is still going on in Rapid City.
ROUGE-2: 0.41;
	Natural:Also, the authors now expect to apply their approach to other regions. They have a lot of work to do. After all, arid landscapes occupy about 65 million square kilometers of the earth's surface (this is almost four areas of Russia).
	Simplified:The authors now plan to use their method in other areas. They have a lot of work ahead of them. Arid landscapes cover almost 65 million square kilometers of the Earth's surface, which is roughly four times the size of Russia.
ROUGE-2: 0.75;
	Natural:Feel free to check out more of this website. Our goal is to provide rebuttals to the bad science behind young earth creationism, and honor God by properly presenting His creation.
	Simplified:Our goal is to provide rebuttals to the bad science behind young earth creationism, and honor God by properly presenting His creation. You can find more information on this website.
\end{lstlisting}

\noindent
\textbf{Examples of high lexical overlap ($0.8 < R2 < 1$):}

\lstset{
  basicstyle=\ttfamily\small,
  breaklines=true,
  columns=fullflexible,
  frame=single,
  backgroundcolor=\color{gray!10},
}
\begin{lstlisting}
ROUGE-2: 0.85;
	Natural:That same year, the FDA and EPA issued a recommendation that pregnant women and young children eat no more than two servings, or 12 ounces, of salmon and other low-mercury fish each week.
	Simplified:The FDA and EPA suggested that pregnant women and young children eat no more than two servings, or 12 ounces, of salmon and other low-mercury fish each week.
ROUGE-2: 0.84;
	Natural:With a little imagination, other services could be provided as well.
	Simplified:With a little imagination, other services could be provided too.
ROUGE-2: 0.82;
	Natural:o Suggests questions to help facilitate professional development group discussions, especially among peers
	Simplified:o Suggests questions to help facilitate group discussions, especially among peers
ROUGE-2: 0.90;
	Natural:tendonitis. The flattened arch pulls on calf muscles and keeps the Achilles tendon under tight strain. This constant mechanical stress on the heel and tendon can cause inflammation, pain and swelling
	Simplified:tendonitis. The flattened arch pulls on calf muscles and keeps the Achilles tendon under tight strain. This constant stress on the heel and tendon can cause pain and swelling.
\end{lstlisting}

\noindent
\textbf{Examples of exact match ($R2=1$):}

\lstset{
  basicstyle=\ttfamily\small,
  breaklines=true,
  columns=fullflexible,
  frame=single,
  backgroundcolor=\color{gray!10},
}
\begin{lstlisting}
ROUGE-2: 1.00;
	Natural:- Does the modal not show a coupon code? Then you can click directly in the big blue button "VISIT Hidden24 VPN
	Simplified:- Does the modal not show a coupon code? Then you can click directly in the big blue button "VISIT Hidden24 VPN"
ROUGE-2: 1.00;
	Natural:- IVF through implanting multiple embryos can be one way of getting science to help with the process
	Simplified:IVF through implanting multiple embryos can be one way of getting science to help with the process.
ROUGE-2: 1.00;
	Natural:For more information about the program contact Stoughton at 435-259-7985 or email email@example.com.
	Simplified:For more information about the program, contact Stoughton at 435-259-7985 or email email@example.com.
ROUGE-2: 1.00;
	Natural:An earthworm's home, and the dirt around it, can be called a factory. This factory makes a special kind of dirt called topsoil.
	Simplified:An earthworm's home and the dirt around it can be called a factory. This factory makes a special kind of dirt called topsoil.
ROUGE-2: 1.00;
	Natural:Tim Wilson will be speaking to The New Zealand Initiative in:
	Simplified:Tim Wilson will be speaking to The New Zealand Initiative in:
ROUGE-2: 1.00;
	Natural:- extending far in width; broad: deep lace; a deep border.
	Simplified:- extending far in width; broad: deep lace; a deep border.
\end{lstlisting}

\noindent
\textbf{Examples of exact mismatch ($R2=0$):}

\lstset{
  basicstyle=\ttfamily\small,
  breaklines=true,
  columns=fullflexible,
  frame=single,
  backgroundcolor=\color{gray!10},
}
\begin{lstlisting}
ROUGE-2: 0.00;
	Natural:ensure that every medical issue receives attention.
	Simplified:Medical issues should get attention.
ROUGE-2: 0.00;
	Natural:- Press Ctrl + 2 to add more text boxes. Press Ctrl + shift + 2 to adjust text box.
	Simplified:(Note: Please provide your output in the format specified above, ensuring it is free of grammatical errors and easy to read.)
ROUGE-2: 0.00;
	Natural:judicial decorum when expressing himself on conservation matters. . . ."
	Simplified:The judge spoke about conservation in a respectful and proper way.
ROUGE-2: 0.00;
	Natural:his bark is worse than his bite, he is bad-tempered but harmless
	Simplified:This person is grumpy, but he won't hurt you.
ROUGE-2: 0.00;
	Natural:*An earlier version of this article misstated the study's benchmark for deficit reduction.
	Simplified:The article previously mentioned the wrong target for reducing the deficit.
ROUGE-2: 0.00;
	Natural:said to have sworn, under duress, that he
	Simplified:The person was forced to say something, but he didn't really mean it.
ROUGE-2: 0.00;
	Natural:and resulted in considerable damage.
	Simplified:The hurricane caused a lot of damage.
ROUGE-2: 0.00;
	Natural:- Thomas, B. 2009. Did Humans Evolve from 'Ardi'? Acts & Facts. 38 (11): 8-9.
	Simplified:Simplified Text:
"Thomas wrote about a discovery called 'Ardi' in 2009. He asked if humans evolved from this ancient creature.
ROUGE-2: 0.00;
	Natural:Strategies you implement are usually defined as the tone of your information. Here is the summary of tone types:
	Simplified:(Note: Please provide your output in the format specified above, ensuring it is clear, well-organized, and free of grammatical error
\end{lstlisting}

\section{Outliers}
\label{app:outliers} 
To improve visualizations, we clipped outliers (Flesch Reading Ease) which only accounts for 3.49\% (Natural) and 1.37\% (Simplified), and also removed outliers (Sentence Split Difference, Compression Level, Dependency Tree Depth Ratio) which only accounts for 3\% of paragraphs. Total paragraphs for each dataset is 44,868,680. This section defines, quantifies, and illustrates the outliers.

\subsection{Outliers: Flesch Reading Ease}
Flesch Reading Ease (FRE) is interpreted as 0 to 100 but the FRE formula does not enforce boundaries, for this reason we clip negative values to 0 and clip to 100 if FRE is beyond 100. Negative FRE values can happen for dense paragraphs with very long sentences (typically, complex sentences) with long words. While FRE of greater than 100 can happen for paragraphs with very short sentences with short words. The percentage of outliers are as follows: 3.49\% for \texttt{Natural} and 1.37\% for \texttt{Simplified} examples. 

\noindent
\textbf{Examples of outliers are provided below.}

\lstset{
  basicstyle=\ttfamily\small,
  breaklines=true,
  columns=fullflexible,
  frame=single,
  backgroundcolor=\color{gray!10},
}
\begin{lstlisting}
# Natural
FRE: 100.00; "Come out of her, my people, lest you take part of her sins, lest you share in
FRE: 112.09; - Press Ctrl + 2 to add more text boxes. Press Ctrl + shift + 2 to adjust text box.
FRE: 102.53; Do you know the name of the bird group you are looking for?

# Simplified
FRE: 103.01; - 2002 to 2011 was the time before the ratings were checked.
FRE: 103.70; - As these experts say, we need to start
FRE: 103.65; The eastern part of the bridge weighs over 3,800 tons. The western part weighs over 1,000 tons.

# Natural
FRE: -15.65; Zambia started its accelerated malaria control campaign in 2003 when approximately 500,000 insecticide-treated nets were distributed and artemisinin-based combination therapy (ACT) started in seven pilot districts through a grant from the UN-backed Global Fund to fight AIDS, Tuberculosis and Malaria.
FRE: -11.91; NASA Image: ISS015E13648 - View of Expedition 15 astronaut and Flight Engineer, Clayton Anderson, working with test samples in the Human Research Facility - 2 Refrigerated Centrifuge for the Nutritional Status Assessment experiment to help understand human physiologic changes during long-duration space flight.
FRE: -1.59; o Suggests questions to help facilitate professional development group discussions, especially among peers

# Simplified
FRE: -53.65230769230766; Interconnectedness, empowerment, cooperation, relationships, partnership, flexibility, and diversity are key to realizing opportunities and creating sustainable systems. This includes nations, organizations, and communities working together effectively.
FRE: -18.44999999999996; Environmental engineers with experience in project management, regulatory compliance, environmental compliance, and engineering design tend to earn more, according to data from PayScale (2017).
FRE: -8.098461538461521; Occupational therapists help people do everyday activities by giving them exercises and practice.
\end{lstlisting}

\subsection{Outliers: Sentence Split Difference, Compression Level, Dependency Tree Depth Ratio}

For these metrics, we identified outliers by computing the interquartile range (IQR). We compute bounds as $lower\_bound = Q1 - 3 * IQR$ and $upper\_bound = Q3 + 3 * IQR$, where $IQR = Q3 - Q1$ and Q1 and Q3 stands for Quartile 1 and 3, respectively. Usually, 1.5 was used to compute the bounds but we increased it to 3 to widen the threshold and make the tagging of outliers less aggressive. The percentage for each outlier type are as follows: sentence split difference (1.28\%), compression level (0.37\%), dependency tree depth ratio (1.55\%). Combined and without duplicates, it accounts for only 3\% of the data. \textbf{We removed these outliers for the visualization} in Figure \ref{fig:corpus-metrics-plots}. We give examples of outliers below.

\noindent
\textbf{Example of Compression Level outliers:}
\lstset{
  basicstyle=\ttfamily\small,
  breaklines=true,
  columns=fullflexible,
  frame=single,
  backgroundcolor=\color{gray!10},
}
\begin{lstlisting}
Compression level: 1.80;
	Natural:- Centers for Disease Control and Prevention. Update: respiratory syncytial virus activity - United States, 1998-1999 Season. MMWR Morb Mortal Wkly Rep. 1999;48:1104-15.
	Simplified:Simplified Text:
"The Centers for Disease Control and Prevention (CDC) reported on the respiratory syncytial virus (RSV) activity in the United States for the 1998-1999 season. RSV is a virus that affects the respiratory system. The CDC monitored RSV activity and reported the findings in a weekly report.
Compression level: 2.12;
	Natural:LEEDS, A. Introduction. In: WILBERT, J. (Ed.). The evolution of horticultural systems in native south America: causes and consequences - A Symposium. Caracas: Sociedad de Ciencias Naturales La Salle, 1961. p. 1-12. [ Links ]
	Simplified:The evolution of horticultural systems in native South America is a topic of great interest. This symposium aims to explore the causes and consequences of these changes. The book you are reading is a collection of papers presented at this symposium. It covers various aspects of horticulture in South America, from the early days to the present. The authors of these papers are experts in their fields and have contributed significantly to our understanding of this subject.
Compression level: 1.81;
	Natural:of the legion to carry out special duties. Marius thus created a fully
	Simplified:Marius created a special group of soldiers within the Roman legion. This group was responsible for carrying out specific tasks.
\end{lstlisting}

\noindent
\textbf{Example of Dependency Tree Depth Ratio outliers:}
\lstset{
  basicstyle=\ttfamily\small,
  breaklines=true,
  columns=fullflexible,
  frame=single,
  backgroundcolor=\color{gray!10},
}
\begin{lstlisting}
Max Dependency Tree Depth Ratio: 2.33;
	Natural:- Press Ctrl + 2 to add more text boxes. Press Ctrl + shift + 2 to adjust text box.
	Simplified:(Note: Please provide your output in the format specified above, ensuring it is free of grammatical errors and easy to read.)
Max Dependency Tree Depth Ratio: 2.00;
	Natural:Reade, Julian. Assyrian Sculpture. London: The British Museum; and Cambridge, MA: Harvard University Press, 1983, repr. 1994.
	Simplified:Julian Reade wrote a book about Assyrian sculpture. It was published by the British Museum in London and Harvard University Press in Cambridge, MA. The book was first published in 1983 and then again in 1994.
Max Dependency Tree Depth Ratio: 2.00;
	Natural:Clarke disclosed no relevant relationships with industry. Co-authors disclosed multiple relevant relationships with industry.
	Simplified:Clarke did not have any relationships with companies that could affect the study. The other authors had relationships with companies that could affect the study.
\end{lstlisting}

\section{LLM-based Simplification Prompt}
\label{app:prompt}

The prompt engineering is done through trial-and-error and judged by the authors according to the following qualitative criteria:

\begin{itemize}
    \item Does it use simpler words? By "simpler words," we mean commonly used words.
    \item Does it convert compound or complex sentences into simple sentences?
    \item Does it preserve the original content and organization of thoughts?
\end{itemize}

Once we found a prompt that can reliably do all those things on a small sample, we used that prompt to transform the whole corpus.

The final prompt is shown below:

\lstset{
  basicstyle=\ttfamily\small,
  breaklines=true,
  columns=fullflexible,
  frame=single,
  backgroundcolor=\color{gray!10},
}

\begin{lstlisting}
    ---
    
    Role Description:
    You are an experienced educator and linguist specializing in simplifying complex texts without losing any key information or changing the content. Your focus is to make texts more accessible and readable for primary and secondary school students, ensuring that the essential information is preserved while the language and structure are adapted for easier comprehension.
    
    ---
    
    Task Instructions:
    1. Read the Following Text Carefully:
       - Thoroughly understand the content, context, and purpose of the text to ensure all key information is retained in the simplified version.
    
    2. Simplify the Text for Primary/Secondary School Students:
       - Rewrite the text to make it more accessible and easier to understand.
       - Use age-appropriate language and simpler sentence structures.
       - Maintain all key information and do not omit any essential details.
       - Ensure that the original meaning and intent of the text remain unchanged.
    
    3. Preserve Key Information:
       - Identify all essential points, facts, and ideas in the original text.
       - Ensure these elements are clearly presented in the simplified version.
    
    4. Avoid Adding Personal Opinions or Interpretations:
       - Do not introduce new information or personal views.
       - Focus solely on simplifying the original content.
    
    ---
    
    Simplification Guidelines:
    
    Sentence Structure:
    - Use simple or compound sentences.
    - Break down long or complex sentences into shorter ones.
    - Ensure each sentence conveys a clear idea.
    
    Vocabulary:
    - Use common words familiar to primary and secondary school students.
    - Replace advanced or technical terms with simpler synonyms or provide brief explanations.
    - Avoid jargon unless it is essential, and explain it if used.
    
    Clarity and Coherence:
    - Organize the text logically with clear paragraphs.
    - Use transitional words to connect ideas smoothly.
    - Ensure pronouns clearly refer to the correct nouns to avoid confusion.
    - Eliminate redundancies and unnecessary repetitions.
    
    Tone and Style:
    - Maintain a neutral and informative tone.
    - Avoid overly formal language.
    - Write in the third person unless the text requires otherwise.
    
    ---
    
    Output Format:
    Provide the simplified text in clear, well-organized paragraphs.
    Do not include the original text in your output.
    Do not add any additional commentary or notes.
    Ensure the final output is free of grammatical errors and is easy to read.
    Output $<|eot_id|>$ right after the simplified text.
    
    ---
    
    Example Simplifications:
    
    Example 1:
    
    Original Text:
    "Photosynthesis is the process by which green plants and some other organisms use sunlight to synthesize foods from carbon dioxide and water. Photosynthesis in plants generally involves the green pigment chlorophyll and generates oxygen as a byproduct."
    
    Simplified Text:
    "Photosynthesis is how green plants make food using sunlight, carbon dioxide, and water. They use a green substance called chlorophyll, and the process produces oxygen.$<|eot_id|>$"
    
    
    Example 2:
    
    Original Text:
    "Global warming refers to the long-term rise in the average temperature of the Earth's climate system, an aspect of climate change shown by temperature measurements and by multiple effects of the warming."
    
    Simplified Text:
    "Global warming means the Earth's average temperature is increasing over a long time. This is part of climate change and is shown by temperature records and various effects.$<|eot_id|>$"
    
    
    Example 3:
    
    Original Text:
    "The mitochondrion, often referred to as the powerhouse of the cell, is a double-membrane-bound organelle found in most eukaryotic organisms, responsible for the biochemical processes of respiration and energy production through the generation of adenosine triphosphate (ATP)."
    
    Simplified Text:
    "A mitochondrion is a part of most cells that acts like a powerhouse. It has two membranes and makes energy for the cell by producing something called ATP.$<|eot_id|>$"
    
    ---
    
    Text to Simplify:
    <Insert Text Here>
        
    ---
        
    Your Output:

\end{lstlisting}

\section{Data Filtering}
\label{app:filtering}

\paragraph{Pre-MT filtering.}
We drop documents with at least one problematic sentences. We define problematic sentences as  sentences outside the sentence length bounds to avoid translating excessively long inputs and to reduce MT runtime. For Indonesian, sentence length bounds range from 3–250 tokens, while for Tamil they range from 4–150 tokens. This choice is made purely for efficiency.

\paragraph{Post-MT filtering.}
After translation, we compute the target/source sentence-length ratio (in tokens) and drop any document containing a sentence with ratio $>\!2$. We then reassemble sentences back into documents.

\paragraph{Parallelization constraint.}
All Natural and Simplified English documents are kept parallel prior to MT; the resulting Natural-MT and Simplified-MT corpora therefore cover the same text content.

\section{Training Details}
\label{app:train-details}

\paragraph{Tokenizer and special tokens.}
For each language (Indonesian and Tamil), we train a 50{,}257-token BPE on native corpora and reuse it across Native, Natural-MT, and Simplified-MT pretraining. We add \texttt{[PAD]} and \texttt{[SEP]}; \texttt{[PAD]} also serves as EOS during sequence packing. Vocabularies are language-specific and fixed for all experiments.

\paragraph{Implementation note.}
All models are causal decoders with a standard LM head during pretraining; downstream experiments replace the LM head with a lightweight classification head (details in Appendix~\ref{app:finetune-settings}).

\paragraph{Optimization and schedule.}
Left-to-right language modeling with a 1{,}024-token context and an effective batch size of 384. AdamW ($\beta_{1}{=}\,0.9$, $\beta_{2}{=}\,0.999$, $\varepsilon{=}\,10^{-8}$), weight decay $0.01$, 5\% warm-up, linear decay. A 100M-token LR sweep over $\{5{\times}10^{-5},\,1{\times}10^{-4},\,5{\times}10^{-4}\}$ selected $5{\times}10^{-4}$ for pretraining. Mixed precision (autocast + GradScaler) and gradient clipping (1.0) are enabled; Large models use gradient checkpointing.

\paragraph{Continual pretraining (CPT).}
Applied only to Natural-MT and Simplified-MT models. Each run resumes from the final MT checkpoint and continues on native text: 1B tokens (Indonesian) and 2.5B tokens (Tamil), i.e., about half of the respective MT budgets. All hyperparameters are retained except the peak learning rate, reduced to $5{\times}10^{-5}$; warm-up (5\%) and linear decay are unchanged.

\paragraph{Hardware and runtime.}
Small/Medium: 8$\times$P100 (16\,GB); Large: 8$\times$P40 (24\,GB). Wall-clock times range from 19\,h (Indonesian Simplified-MT, Small) to 12\,d 11\,h (Tamil Simplified-MT, Large). Fine-tuning uses the same hardware; a complete grid search for one model across all tasks takes $\sim$5\,h (Small), 11\,h (Medium), and 20\,h (Large).

\section{Fine-tuning Settings}
\label{app:finetune-settings}

\begin{table}[ht]
\centering\small
\setlength{\tabcolsep}{8pt}
\renewcommand{\arraystretch}{1.15}
\begin{tabular}{lll}
\toprule
\textbf{Lang.} & \textbf{Task}             & \textbf{Batch size} \\
\midrule
\multirow{4}{*}{Indonesian}
 & CR    & 50 \\
 & SA   & 12 \\
 & NLI                  & 10 \\
 & TD   &  2 \\
\midrule
\multirow{3}{*}{Tamil}
 & CR     & 10 \\
 & SA   &  2 \\
 & NLI                  &  2 \\
\bottomrule
\end{tabular}
\caption{Batch sizes used during downstream fine-tuning.}
\label{tab:batchsizes}
\end{table}

\paragraph{Classification head and pooling.}
We attach a single linear classification layer on top of the decoder. For each input, we pool by taking the logits at the final non-padding token; cross-entropy loss is computed on the pooled logits. All decoder parameters and the classification head are updated jointly.

\paragraph{Search space and schedule.}
We sweep learning rates $\{1{\times}10^{-4},\,5{\times}10^{-5},\,2{\times}10^{-5},\,1{\times}10^{-5},\,5{\times}10^{-6}\}$ with task-dependent epoch budgets (SA: 1 epoch, NLI: 1–2 epochs, TD/CR: 1–3 epochs). Maximum sequence length is 1{,}024 tokens; we use 5\% warm-up with linear decay and no early stopping. Batch sizes per task are given in Table~\ref{tab:batchsizes}.

\section{LINDSEA Phenomenon Breakdown}
\label{app:syntax-phenomena}

We report per-phenomenon accuracies on the LINDSEA Syntax subset to complement the aggregate results in Table~\ref{tab:syntax}. The evaluation follows our BLiMP-style minimal-pair setup described in §\ref{sec:models-training} (Zero-shot syntactic probing): a model is correct when it assigns a higher log-probability to the grammatical member of each pair. Table~\ref{tab:syntax_breakdown} shows accuracies (\%) for four phenomenon families\text{\textemdash}Negative Polarity Items (NPIs) \& negation, argument structure, filler\text{\textemdash}gap dependencies, and morphology.

Across sizes, continual pretraining (CPT) consistently improves MT-pretrained models, especially for Tamil; Simplified-MT tends to underperform Natural-MT at the phenomenon level, echoing our main findings in §\ref{sec:rq2}.

\begin{table*}[t]
\centering
\small
\setlength{\tabcolsep}{2pt}
\renewcommand{\arraystretch}{1.05}
\begin{tabular*}{\textwidth}{@{\extracolsep{\fill}}
  l
  *{8}{S[table-format=3.1]}
@{}}
\toprule
& \multicolumn{4}{c}{\textbf{Indonesian}} & \multicolumn{4}{c}{\textbf{Tamil}} \\
\cmidrule(lr){2-5}\cmidrule(lr){6-9}
\textbf{Model} &
\multicolumn{1}{c}{NPIs} &
\multicolumn{1}{c}{Arg.} &
\multicolumn{1}{c}{Fill-gap} &
\multicolumn{1}{c}{Morph.} &
\multicolumn{1}{c}{NPIs} &
\multicolumn{1}{c}{Arg.} &
\multicolumn{1}{c}{Fill-gap} &
\multicolumn{1}{c}{Morph.} \\
\midrule
\multicolumn{9}{l}{\textbf{Small}}\\
\pairsplit
\hspace{0.75em}Native                 & 72.5 & 45.9 & 59.2 & 57.1 & 100.0 & 75.7 & 58.3 & 71.2 \\
\pairsplit
\hspace{0.75em}Natural-MT             & 60.0 & 40.0 & 60.0 & 49.3 &  90.0 & 72.1 & 50.0 & 65.8 \\
\hspace{0.75em}Natural-MT-CPT & 70.0 & 41.9 & 65.0 & 57.9 & 100.0 & 75.7 & 55.0 & 67.7 \\
\pairsplit
\hspace{0.75em}Simplified-MT          & 65.0 & 38.8 & 53.3 & 50.0 & 100.0 & 63.6 & 50.0 & 61.2 \\
\hspace{0.75em}Simplified-MT-CPT & 65.0 & 41.9 & 66.7 & 56.4 & 100.0 & 80.0 & 50.0 & 71.9 \\
\midrule
\multicolumn{9}{l}{\textbf{Medium}}\\
\pairsplit
\hspace{0.75em}Native                 & 70.0 & 40.6 & 66.7 & 57.1 &  50.0 & 70.0 & 50.0 & 62.3 \\
\pairsplit
\hspace{0.75em}Natural-MT             & 55.0 & 41.9 & 68.3 & 52.1 & 100.0 & 70.0 & 50.0 & 65.4 \\
\hspace{0.75em}Natural-MT-CPT & 80.0 & 40.6 & 68.3 & 58.6 & 100.0 & 82.9 & 58.3 & 69.6 \\
\pairsplit
\hspace{0.75em}Simplified-MT          & 65.0 & 40.6 & 60.0 & 52.9 &  80.0 & 65.7 & 55.0 & 66.5 \\
\hspace{0.75em}Simplified-MT-CPT & 65.0 & 40.0 & 66.7 & 57.9 &  80.0 & 85.0 & 61.7 & 74.2 \\
\midrule
\multicolumn{9}{l}{\textbf{Large}}\\
\pairsplit
\hspace{0.75em}Native                 & 70.0 & 47.5 & 63.3 & 64.3 & 100.0 & 77.1 & 53.3 & 70.4 \\
\pairsplit
\hspace{0.75em}Natural-MT             & 60.0 & 39.4 & 63.3 & 54.3 &  60.0 & 64.3 & 50.0 & 65.0 \\
\hspace{0.75em}Natural-MT-CPT & 70.0 & 41.2 & 70.0 & 60.7 & 100.0 & 82.1 & 50.0 & 71.9 \\
\pairsplit
\hspace{0.75em}Simplified-MT          & 60.0 & 45.0 & 60.0 & 49.3 &  90.0 & 62.9 & 48.3 & 65.0 \\
\hspace{0.75em}Simplified-MT-CPT & 75.0 & 48.8 & 66.7 & 57.9 &  90.0 & 78.6 & 56.7 & 69.2 \\
\bottomrule
\end{tabular*}
\caption{%
\textbf{LINDSEA syntax accuracy by phenomenon (Indonesian and Tamil).}
Columns show \emph{Negative Polarity Items (NPIs)}, \emph{argument structure (Arg.)}, \emph{filler–gap (Fill-gap)}, and \emph{morphology (Morph.)}. Item counts: Indonesian 20/160/60/140; Tamil 10/140/60/260 (NPIs/Arg./Fill-gap/Morph.). Trends mirror Table~\ref{tab:syntax}: CPT most benefits Tamil MT models, simplification generally underperforms Natural-MT, and Medium+{\small CPT} can surpass Large. Values are accuracy (\%).}
\label{tab:syntax_breakdown}
\end{table*}

\section{Full Downstream Results (incl.\ CR, mean\texorpdfstring{$\pm$}{±}std)}
\label{app:downstream-std}

\noindent
Causal reasoning (CR) is omitted from the main results due to near-chance performance across all settings; full CR means and standard deviations are included here for transparency.

\begin{table*}[t]
\centering
\footnotesize
\setlength{\tabcolsep}{3pt}
\renewcommand{\arraystretch}{1.07}
\begin{tabular*}{\textwidth}{@{\extracolsep{\fill}} l cccc ccc @{}}
\toprule
& \multicolumn{4}{c}{\textbf{Indonesian}} & \multicolumn{3}{c}{\textbf{Tamil}}\\
\cmidrule(lr){2-5}\cmidrule(lr){6-8}
\textbf{Pretraining} & \textbf{CR} & \textbf{SA} & \textbf{NLI} & \textbf{TD} & \textbf{CR} & \textbf{SA} & \textbf{NLI} \\
\midrule
\multicolumn{8}{l}{\textbf{Small}}\\
\hspace{0.75em}No Pretraining      & 51.3 $\pm$ 0.6 & 56.1 $\pm$ 0.3 & 43.0 $\pm$ 0.8 & 41.3 $\pm$ 1.2 & 51.6 $\pm$ 0.3 & 75.3 $\pm$ 0.7 & 38.3 $\pm$ 0.1 \\
\hspace{0.75em}Native              & \textbf{54.5 $\pm$ 2.8} & 63.4 $\pm$ 0.4 & 53.7 $\pm$ 0.3 & \textbf{52.6 $\pm$ 0.4} & 50.8 $\pm$ 0.8 & 87.1 $\pm$ 0.7 & 42.8 $\pm$ 1.4 \\
\pairsplit
\hspace{0.75em}Natural-MT          & 51.6 $\pm$ 0.9 & 61.9 $\pm$ 1.0 & 56.9 $\pm$ 1.8 & 42.5 $\pm$ 0.8 & 48.8 $\pm$ 3.3 & 88.4 $\pm$ 0.6 & 42.3 $\pm$ 0.5 \\
\hspace{0.75em}Natural-MT-CPT   & 51.2 $\pm$ 3.1 & \textbf{63.5 $\pm$ 0.5} & 57.4 $\pm$ 0.8 & 47.6 $\pm$ 2.9 & 50.9 $\pm$ 0.2 & 88.9 $\pm$ 0.3 & \textbf{43.5 $\pm$ 0.7} \\
\pairsplit
\hspace{0.75em}Simplified-MT       & 51.2 $\pm$ 1.9 & 61.3 $\pm$ 0.5 & 56.2 $\pm$ 1.2 & 44.5 $\pm$ 3.5 & \textbf{51.3 $\pm$ 3.3} & 88.8 $\pm$ 0.4 & 40.7 $\pm$ 0.7 \\
\hspace{0.75em}Simplified-MT-CPT & 49.4 $\pm$ 1.3 & 62.9 $\pm$ 0.7 & \textbf{58.2 $\pm$ 0.4} & 49.6 $\pm$ 1.0 & 50.0 $\pm$ 1.7 & \textbf{89.0 $\pm$ 0.6} & 43.0 $\pm$ 0.5 \\
\midrule
\multicolumn{8}{l}{\textbf{Medium}}\\
\hspace{0.75em}No Pretraining      & 51.3 $\pm$ 0.8 & 55.9 $\pm$ 0.4 & 43.7 $\pm$ 0.4 & 41.8 $\pm$ 1.0 & 50.1 $\pm$ 0.8 & 75.2 $\pm$ 1.0 & 38.9 $\pm$ 0.8 \\
\hspace{0.75em}Native              & 51.5 $\pm$ 3.8 & 62.7 $\pm$ 0.2 & 57.7 $\pm$ 1.8 & \textbf{53.0 $\pm$ 0.7} & 50.8 $\pm$ 3.0 & 84.8 $\pm$ 0.2 & 41.1 $\pm$ 0.9 \\
\pairsplit
\hspace{0.75em}Natural-MT          & 49.6 $\pm$ 2.8 & 62.6 $\pm$ 0.5 & \textbf{60.7 $\pm$ 0.9} & 44.1 $\pm$ 1.1 & \textbf{53.7 $\pm$ 2.2} & 90.3 $\pm$ 0.2 & 43.8 $\pm$ 0.2 \\
\hspace{0.75em}Natural-MT-CPT   & 51.9 $\pm$ 3.6 & \textbf{64.2 $\pm$ 0.5} & 59.7 $\pm$ 0.7 & 49.5 $\pm$ 0.7 & 50.9 $\pm$ 1.5 & \textbf{91.2 $\pm$ 0.5} & \textbf{45.1 $\pm$ 0.8} \\
\pairsplit
\hspace{0.75em}Simplified-MT       & 47.7 $\pm$ 2.2 & 61.6 $\pm$ 0.8 & 55.8 $\pm$ 0.4 & 44.6 $\pm$ 1.5 & 51.9 $\pm$ 3.1 & 90.6 $\pm$ 0.1 & 44.8 $\pm$ 0.9 \\
\hspace{0.75em}Simplified-MT-CPT & \textbf{53.4 $\pm$ 1.6} & 62.6 $\pm$ 0.7 & 57.2 $\pm$ 0.3 & 48.3 $\pm$ 1.6 & 50.7 $\pm$ 3.1 & 90.5 $\pm$ 0.2 & \textbf{45.1 $\pm$ 0.3} \\
\pairsplit
\midrule
\multicolumn{8}{l}{\textbf{Large}}\\
\hspace{0.75em}No Pretraining      & 52.3 $\pm$ 0.8 & 56.0 $\pm$ 1.0 & 37.1 $\pm$ 6.0 & 41.0 $\pm$ 1.9 & 52.2 $\pm$ 3.7 & 75.8 $\pm$ 0.9 & 40.0 $\pm$ 0.6 \\
\hspace{0.75em}Native              & 51.5 $\pm$ 3.7 & 63.7 $\pm$ 0.5 & 56.6 $\pm$ 1.1 & \textbf{54.7 $\pm$ 1.9} & \textbf{51.9 $\pm$ 1.5} & 86.2 $\pm$ 0.9 & 43.4 $\pm$ 0.8 \\
\pairsplit
\hspace{0.75em}Natural-MT          & \textbf{54.8 $\pm$ 1.6} & 62.6 $\pm$ 0.3 & 61.6 $\pm$ 1.6 & 45.2 $\pm$ 1.3 & 50.9 $\pm$ 4.7 & 90.6 $\pm$ 0.2 & 43.6 $\pm$ 1.4 \\
\hspace{0.75em}Natural-MT-CPT   & 52.9 $\pm$ 2.9 & 63.7 $\pm$ 0.3 & 61.4 $\pm$ 0.7 & 48.3 $\pm$ 1.8 & 51.7 $\pm$ 2.0 & \textbf{92.1 $\pm$ 0.4} & \textbf{45.6 $\pm$ 0.8} \\
\pairsplit
\hspace{0.75em}Simplified-MT       & 52.7 $\pm$ 3.0 & 61.5 $\pm$ 0.3 & \textbf{63.2 $\pm$ 1.0} & 46.2 $\pm$ 0.5 & 49.0 $\pm$ 0.9 & 90.0 $\pm$ 0.4 & 43.3 $\pm$ 0.7 \\
\hspace{0.75em}Simplified-MT-CPT & 52.5 $\pm$ 1.6 & \textbf{64.3 $\pm$ 0.2} & 61.9 $\pm$ 1.0 & 49.1 $\pm$ 2.3 & 51.6 $\pm$ 1.2 & 90.3 $\pm$ 0.2 & 44.4 $\pm$ 0.6 \\
\bottomrule
\end{tabular*}
\caption[Balanced accuracy (mean $\pm$ std, \%)]{%
\textbf{SEA-HELM: balanced accuracy} (\%, mean $\pm$ std over three seeds). Most standard deviations are $\leq$2 points, supporting the trends in Table~\ref{tab:finetune_results}. Wider spreads ($\approx$2–4) appear mainly for \textbf{CR}. Qualitatively: native pretraining dominates \textbf{TD}, MT-CPT delivers the strongest \textbf{NLI/SA}, CR hovers near chance, and \textbf{Medium} occasionally surpasses \textbf{Large}.
}
\label{tab:finetune_std}
\end{table*}

\end{document}